\documentclass[conference]{IEEEtran}
\IEEEoverridecommandlockouts
\usepackage{cite}
\usepackage{amsmath,amssymb,amsfonts}
\usepackage{algorithmic}
\usepackage{textcomp}
\usepackage{xcolor}
\usepackage{booktabs}
\usepackage{multirow}
\usepackage{colortbl}
\usepackage{xcolor}
\usepackage{graphicx}
\usepackage{orcidlink}
\usepackage{comment}
\usepackage{subcaption}
\usepackage{dblfloatfix}   
\def\BibTeX{{\rm B\kern-.05em{\sc i\kern-.025em b}\kern-.08em
    T\kern-.1667em\lower.7ex\hbox{E}\kern-.125emX}}
\definecolor{best}{RGB}{220,80,80}
\definecolor{second}{RGB}{80,120,200}

\begin{document}

\title{FMMVCC: Fuzzy Mamba-based Multi-View Contrastive Clustering for Univariate Time Series\\
}

\author{\IEEEauthorblockN{1\textsuperscript{st} Donato Cerciello* \orcidlink{0009-0004-8517-5519}}
\IEEEauthorblockA{\textit{Departamento de Sistemas Informáticos,} \\
\textit{Universidad Politécnica de Madrid}\\
Madrid, Spain \\
\href{mailto:donato.cerciello@upm.es}{donato.cerciello@upm.es}}
\and
\IEEEauthorblockN{2\textsuperscript{nd} Leonardo Schiavo \orcidlink{0009-0007-5023-0946}}
\IEEEauthorblockA{\textit{Departamento de Sistemas Informáticos,} \\
\textit{Universidad Politécnica de Madrid}\\
Madrid, Spain \\
\href{mailto:leonardo.schiavo@upm.es}{leonardo.schiavo@upm.es}}
\and
\IEEEauthorblockN{3\textsuperscript{rd} Angel Panizo-LLedot \orcidlink{0000-0002-2195-3527}}
\IEEEauthorblockA{\textit{Departamento de Sistemas Informáticos,} \\
\textit{Universidad Politécnica de Madrid}\\
Madrid, Spain \\
\href{mailto:angel.panizo@upm.es}{angel.panizo@upm.es}}
\and
\IEEEauthorblockN{4\textsuperscript{th} Javier Huertas Tato \orcidlink{0000-0003-4127-5505}}
\IEEEauthorblockA{\textit{Departamento de Sistemas Informáticos,} \\
\textit{Universidad Politécnica de Madrid}\\
Madrid, Spain \\
\href{mailto:javier.huertas.tato@upm.es}{javier.huertas.tato@upm.es}}
\and
\IEEEauthorblockN{5\textsuperscript{th} David Camacho \orcidlink{0000-0002-5051-3475}}
\IEEEauthorblockA{\textit{Departamento de Sistemas Informáticos,} \\
\textit{Universidad Politécnica de Madrid}\\
Madrid, Spain \\
\href{mailto:david.camacho@upm.es}{david.camacho@upm.es}}
\thanks{*Corresponding author: Donato Cerciello}
}

\maketitle

\begin{abstract}
In many realistic scenarios, large volumes of time series data are generated with limited or expensive annotations. This limitation makes supervised learning methods difficult to apply and leads to the use of unsupervised approaches capable of discovering meaningful structures directly from raw data. Clustering therefore plays a crucial role in organizing time series into groups that share similar temporal patterns, enabling exploratory analysis and downstream tasks without requiring manual labeling. However, existing deep clustering methods often struggle to capture long-range temporal dependencies or rely on architectures with high computational cost. This paper introduces FMMVCC, a Mamba-based deep clustering framework for time series that leverages state space sequence modeling to efficiently learn temporal representations with linear complexity. Additionally, it utilizes multi-view self-supervised learning with temporal masking and augmentations. Experimental evaluation in 15 benchmark datasets proves that FMMVCC consistently outperforms state-of-the-art baselines, achieving the best overall performance in 29 of 60 total metric evaluations and the highest average rank in all tested scenarios.
\end{abstract}


\section{Introduction}
The rapid expansion of devices and sensor networks on the Internet of Things (IoT) has led to an explosion of time series data in domains such as smart cities, industrial systems, healthcare and environmental monitoring \cite{schlegel2025towards}. These continuous streams of temporal data contain valuable information for understanding patterns, detecting anomalies, and monitoring system behavior. However, obtaining labeled data in such environments is often expensive, time-consuming, or even infeasible due to the scale and complexity of sensor deployments. As a result, unsupervised learning methods, and in particular time series clustering, have become increasingly important for discovering latent structures in sequential data \cite{SelfSupervisedLearningforTimeSeriesAnalysis}. \\
Time series clustering is defined as an unsupervised learning method that groups similar time series patterns without any supervision. Traditional clustering methods, such as K-means and hierarchical clustering, have been shown to usually require handcrafted measures of similarity, such as Euclidean distance and DTW \cite{paparrizos2024bridging}. Although these methods have been shown to be successful in some cases, they have some difficulties in dealing with the complex temporal relationships and nonlinear dynamics that can be observed in recent high-dimensional time series data. For that reason, recent research has proposed deep learning-based clustering methods. \\
In recent years, contrastive learning (CL) has been established as an effective framework for the self-supervised representation learning method. By maximizing the agreement between different views of the same data and minimizing the agreement between different samples, CL has the ability to learn discriminative representations without the need for any supervision \cite{meng2023unsupervised}. Several works have successfully applied CL to time series analysis by generating multiple augmented views of a signal through transformations. This enables the model to capture invariant representations that preserve the underlying temporal structure \cite{emtc}. However, the design of effective multi-view strategies and the incorporation of clustering objectives is an area of research that needs to be addressed. \\
Another key challenge in time series modeling is capturing long-range temporal dependencies efficiently. Traditional recurrent architectures have been widely used for sequential modeling, but they often struggle with very long sequences and limited parallelization \cite{somvanshi2025s4}. More recently, transformer-based architectures have demonstrated strong performance in sequence modeling tasks \cite{wen2022transformers}, yet their quadratic complexity with respect to sequence length can become computationally expensive in large-scale IoT environments. To address these limitations, state space models (SSMs) have recently re-emerged as an efficient alternative for sequence modeling. In particular, the Mamba architecture introduces a selective state space mechanism that enables linear-time sequence processing while maintaining strong modeling capacity for long temporal contexts \cite{gu2024mamba}. Despite its promising performance in sequence modeling tasks, its potential for unsupervised time series clustering remains largely unexplored \cite{somvanshi2025s4}. \\
To address these challenges, we propose Fuzzy Mamba-based Multi-View Contrastive Clustering (FMMVCC), a deep clustering framework for time series data. The proposed approach employs Mamba-based encoders to effectively capture long-range temporal dependencies while maintaining computational efficiency. To improve representation learning, multiple views of each time series are generated through temporal masking and augmentation strategies, enabling contrastive self-supervised training. In addition, a fuzzy clustering objective is introduced to produce soft cluster assignments, allowing the model to better handle ambiguous temporal patterns and improve cluster separability in the learned latent space. 

\subsection{Contributions}
The main contributions of this work can be summarized as follows:
\begin{itemize}
\item \textbf{Mamba-based Representation Learning for Time Series Clustering.} 
A deep clustering framework is introduced that leverages Mamba-based encoders to model long-range temporal dependencies in time series data while maintaining linear-time computational complexity. This architecture provides an efficient alternative to recurrent and transformer-based models for large-scale sequential data.
\item \textbf{Multi-View Masked Representation Learning.} 
Multiple views of each time series are generated through stochastic temporal augmentations combined with random temporal masking. The resulting views provide diverse but consistent observations of the same sequence, allowing the model to learn robust representations under both temporal perturbations and missing data.
\item \textbf{Cluster-Oriented Representation Learning.} 
A clustering objective is integrated into the representation learning process to progressively structure the latent space and improve the separability of the cluster for time series data.
\end{itemize}
The paper is organized as follows. Section \ref{sec:related}  reviews related work on deep clustering and time series representation learning. Section \ref{sec:methodology} describes the proposed framework. Section \ref{sec:results} presents the experimental results. Lastly, section \ref{sec:conclusions} discusses concluding remarks.

\section{Related Works}\label{sec:related}
\subsection{Contrastive Learning for Time Series Clustering}
CL has emerged as a powerful paradigm for self-supervised representation learning in time series, which enables the model to extract discriminative features without requiring any labels. The core idea is to maximize the agreement between different views of the same sample while minimizing similarity with other samples \cite{meng2023unsupervised}. A significant number of recent studies show that the effectiveness of contrastive learning is largely dependent on the quality and variety of the views. \\
Current methods, such as TS2Vec \cite{yue2022ts2vec}, make use of hierarchical temporal contrasting to extract both local and global temporal dependencies. Similarly, many methods make use of data augmentation methods such as scaling, jittering, and permutation to make the model invariant to common distortions in time series data \cite{emtc}. However, these methods are mainly based on deterministic data enhancement and may not be effective in capturing uncertainty caused by missing data. \\
To address uncertainty in clustering, recent work has combined contrastive learning with fuzzy clustering. In particular, FCACC \cite{FCACC} introduces a cluster-aware contrastive framework with soft assignments, improving robustness in ambiguous time series scenarios. \\
In addition, the concept of temporal masking has also been explored as an effective approach to generate informative and challenging views. In these methods, the model is not required to reconstruct the input data, but to make it invariant to masking. This provides an opportunity for the model to extract robust and invariant representations even in incomplete data. In this direction, the EMTC \cite{emtc} framework adapts the concept of evolving masking to effectively detect and reduce the relevance of less important timestamp. However, such approaches are typically built on attention-based architectures, which suffer from quadratic complexity and limited scalability for long sequences \cite{wen2022transformers}. \\
In contrast to existing approaches, the proposed framework combines contrastive learning with stochastic multi-view masking and cross-view reconstruction, enabling robust representation learning under missing data. It leverages the causal and selective dynamics of Mamba, allowing the model to capture temporal dependencies while filtering irrelevant information in a scalable manner. 

\subsection{Mamba-based Sequence Modeling for Time Series}

Recently, State Space Models (SSMs) have been rediscovered as a promising solution, providing a linear complexity solution along with robust sequence modeling. Mamba provides a selective state space mechanism which filters relevant information based on input-dependent transitions. This enables efficient representation \cite{gu2024mamba}. \\
Recently, a study on PG-Mamba \cite{pgmamba} was conducted to explore the use of Mamba-based clustering frameworks by incorporating a graph-based structure to capture global relationships between time series. Although such frameworks provide promising results, robustness is not a key focus of such frameworks. \\
In the presented framework, Mamba has been used within a multi-view contrastive setting to provide robustness via random masking and augmentation. A fuzzy objective is used to provide clustering. This enables efficient representation of long-range relationships within a representation that is invariant and discriminative.

\section{Methodology}\label{sec:methodology}
\begin{figure}[!t]
    \centering
    \includegraphics[width=\columnwidth]{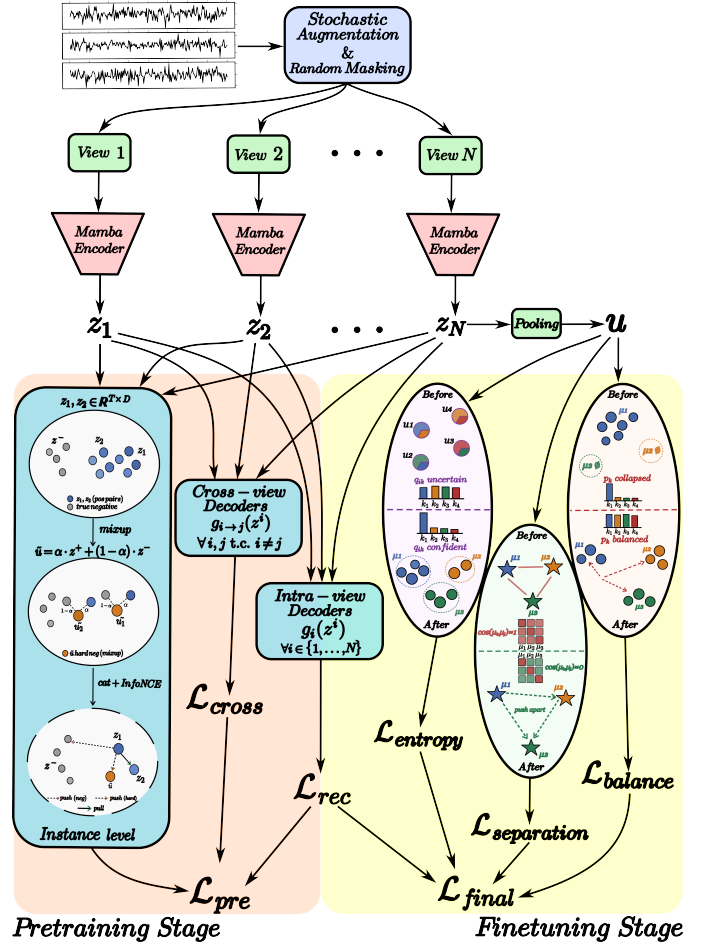}
    \caption{Overview of the proposed FMMVCC framework. Multiple masked views of the input time series are encoded via Mamba-based encoders. During pretraining, intra-view and cross-view decoders enforce reconstruction and consistency, and a contrastive objective aligns representations across views. In the finetuning stage, the fused representations are optimized with a clustering objective to produce soft cluster assignments.}
    \label{fig:framework}
\end{figure}

A complete overview of the FMMVCC framework is presented
 in Figure \ref{fig:framework}.

\subsection{Multi-view Generation via Temporal Masking}

Given an input time series $x \in \mathbb{R}^{T \times D}$, where $T$ denotes the sequence length and $D$ the number of variables, multiple views of the same signal are generated by stochastic augmentation and temporal masking. This strategy produces partially observed and perturbed versions of the original signal, simulating realistic conditions such as noisy or missing sensor measurements in IoT environments.

For each input time series, $N$ views are generated as
\[
x^{(v)} = M^{(v)} \odot A^{(v)}(x), \quad v = 1,\dots,N
\]

where $\odot$ denotes element-wise multiplication, $A^{(v)}(\cdot)$ is the stochastic augmentation operator, and $M^{(v)}(\cdot) \in \{ 0, 1\}^{T \times D}$ represents a binary observation mask operator.\\
The augmentation operator $A^{(v)}$ is sampled from a predefined set of transformations, ensuring that different views receive distinct perturbations. Specifically, stochasticity is introduced through Gaussian noise for jittering, random feature-wise factors for scaling, and random segment reshuffling for permutation. \\
The masking operator $M^{(v)}$ simulates realistic missingness patterns by combining structured and unstructured drops for a target rate $\rho$. Contiguous temporal drops are generated by removing random segments up to a maximal length $L_{max}$, while scattered point-wise drops are sampled uniformly from the remaining positions. To further increase view diversity, $M^{(v)}$ is obtained by equally mixing two independently sampled masks.
\\ 
Each resulting masked view is processed independently by a dedicated encoder. This multi-view generation strategy encourages the model to learn representations that are robust to noise, missing observations, and temporal perturbations.

\subsection{Mamba-Based Representation Learning}

Each masked view is processed by a Mamba-based encoder that maps the input sequence into a latent representation space. Given a masked sequence $x^{(v)}$, the encoder produces a latent sequence representation

\[
z^{(v)} = f_{\theta_v}(x^{(v)}), \quad z^{(v)} \in \mathbb{R}^{T \times d}
\]

where $d$ denotes the latent dimensionality. The encoder consists of an input projection layer followed by a stack of Mamba blocks and a linear output projection. Each Mamba block implements a selective state space model capable of modeling long-range temporal dependencies while maintaining linear-time computational complexity.

\subsection{Cross-View and Intra-View Reconstruction}

To enforce representation consistency, two types of decoder networks are introduced: intra-view decoders and cross-view decoders.

The intra-view decoder reconstructs the original input sequence from its latent representation:

\[
\hat{x}^{(v)} = g_v(z^{(v)})
\]

where $g_v(\cdot)$ denotes the decoder associated with view $v$. The reconstruction loss is defined as the mean squared error (MSE):

\[
\mathcal{L}_{rec} =
\frac{1}{N}\sum_{v=1}^{N}
\| x - \hat{x}^{(v)} \|_2^2
\]

In addition, cross-view decoders enforce consistency between representations of different views. For two views $i$ and $j$, the latent representation from one view is used to reconstruct the latent representation of the other view:

\[
\tilde{z}^{(j)} = g_{i \rightarrow j}(z^{(i)})
\]

The cross-view reconstruction loss is defined as

\[
\mathcal{L}_{cross} =
\frac{1}{N(N-1)}
\sum_{i \neq j}
\| z^{(j)} - \tilde{z}^{(j)} \|_2^2
\]

To further align the representations obtained from different views, an instance-level contrastive objective is applied to the reconstructed latent representations. Given two reconstructed representations $\tilde{z}^{(i)}$ and $\tilde{z}^{(j)}$ corresponding to the same time series, the model maximizes their similarity while contrasting them against representations of other samples in the batch.

Formally, the contrastive loss is defined as

\[
\mathcal{L}_{contrast} =
-
\log
\frac{\exp(\text{sim}(\tilde{z}_i,\tilde{z}_j)/\tau)}
{\sum_{k=1}^{N}
\exp(\text{sim}(\tilde{z}_i,\tilde{z}_k)/\tau)}
\]

where $\text{sim}(\cdot,\cdot)$ denotes a similarity function, $\tau$ is a temperature parameter, and $N$ denotes the number of samples in the batch. Additionally, a mixup-based strategy is employed to generate harder negative samples by combining positive and negative representations, improving the discriminative capability of the learned latent space.

The overall pretraining objective combines all the previous losses:

\[
\mathcal{L}_{pre} =
\lambda_1 \mathcal{L}_{contrast}
+
\lambda_2 \mathcal{L}_{cross}
+
\lambda_3 \mathcal{L}_{rec}
\]

where $\lambda_1$, $\lambda_2$, and $\lambda_3$ control the contribution of each component.

\subsection{Clustering-Oriented Finetuning}

After the pretraining stage, the model is finetuned using a clustering-oriented objective
designed to structure the latent space and improve cluster separability.

Given the latent sequence representations $\{z^{(v)}\}_{v=1}^{V}$, a temporal pooling
mechanism aggregates each sequence into a fixed-length vector, which is then averaged
across views to obtain a single representation $u \in \mathbb{R}^d$ per sample.

Soft cluster assignments are computed through a cosine-similarity-based softmax over a
set of learnable cluster prototypes $\{\mu_k\}_{k=1}^{K}$:

\begin{equation*}
        q_{ik} =
        \frac{\exp\!\left(\langle \bar{u}_i,\, \bar{\mu}_k \rangle / \tau\right)}
        {\sum_{j=1}^{K}\exp\!\left(\langle \bar{u}_i,\, \bar{\mu}_j \rangle / \tau\right)}
\end{equation*}

where $\bar{u}_i$ and $\bar{\mu}_k$ denote $L_2$-normalized representations and
prototypes, and $\tau$ is a temperature parameter.

The clustering objective combines three complementary terms:

\begin{equation*}
    \begin{cases}
        \mathcal{L}_{entropy} =
            \dfrac{1}{N}\sum_{i,k} q_{ik}\log q_{ik} \\[8pt]
        \mathcal{L}_{balance} =
            \sum_{k} p_k \log p_k, \quad p_k = \tfrac{1}{N}\sum_{i} q_{ik} \\[8pt]
        \mathcal{L}_{separation} =
            \dfrac{1}{K(K-1)}\sum_{i \neq j}\!\left(\bar{\mu}_i^\top \bar{\mu}_j\right)^2
    \end{cases}
\end{equation*}

The entropy term encourages confident assignments, the balance term mitigates cluster
collapse, and the separation term promotes diversity among prototypes, yielding the
clustering objective

\begin{equation*}
    \mathcal{L}_{cluster} =
    \mathcal{L}_{entropy} +
    \lambda_b\, \mathcal{L}_{balance} +
    \lambda_s\, \mathcal{L}_{separation}
\end{equation*}

where $\lambda_b$ and $\lambda_s$ are respectively the balance and the separation weights. The overall finetuning objective is therefore defined as

\begin{equation*}
    \mathcal{L}_{final} =
    \lambda_r\, \mathcal{L}_{rec}
    + \lambda_c\, \mathcal{L}_{cluster}
\end{equation*}
where $\lambda_r$ and $\lambda_c$ control the contribution of each component.

\begin{table}[t]
  \centering
  \caption{UCR Univariate Time-Series Datasets Used in Experiments}
  \label{tab:datasets}
  \begin{tabular}{l|c|c|c}
    \toprule
    \textbf{Dataset} & \textbf{Samples} & \textbf{Length} & \textbf{Classes} \\
    \midrule
    ACSF1 & 100 & 1460 & 10 \\
    Adiac & 390 & 176 & 37 \\
    AllGestureWiimoteX & 300 & 500 & 10 \\
    AllGestureWiimoteZ & 300 & 500 & 10 \\
    BirdChicken & 20 & 512 & 2 \\
    Car & 60 & 577 & 4 \\
    CricketX & 390 & 300 & 12 \\
    CricketZ & 390 & 300 & 12 \\
    Crop & 7200 & 46 & 24 \\
    DistalPhalanx & 400 & 80 & 3 \\
    ECG200 & 100 & 96 & 2 \\
    ElectricDevices & 8926 & 96 & 7 \\
    FaceAll & 560 & 131 & 14 \\
    FreezerRegularTrain & 150 & 301 & 2 \\
    SyntheticControl & 300 & 60 & 6 \\
    \bottomrule
  \end{tabular}
\end{table}

\section{Experimental Results}\label{sec:results}
\subsection{Dataset}
The proposed method is evaluated on datasets from the UCR Time Series Archive \cite{dau2019ucr}, a widely used benchmark in the time series data mining community. The archive contains a large collection of univariate time series datasets spanning various domains. As described in Table \ref{tab:datasets}, a subset of representative datasets has been selected to assess the performance of clustering under varying sequence lengths and data characteristics.

\begin{table}[t]
\centering
\caption{Clustering performance comparison across multiple datasets. Best results are highlighted in red and second-best in blue.}
\scriptsize
\setlength{\tabcolsep}{3pt}
\renewcommand{\arraystretch}{0.9}
\resizebox{\columnwidth}{!}{%
\begin{tabular}{llccccc|c}
\toprule
 & & \multicolumn{5}{c}{Baselines} & \multicolumn{1}{c}{Proposed} \\
\cmidrule(lr){3-7} \cmidrule(lr){8-8}
Datasets & Metrics & TSK-DTW & HDBSCAN & PG-Mamba & FCACC & EMTC & \textbf{FMMVCC} \\
\midrule
\multirow{4}{*}{ACSF1} & F1 & 0.0621 & 0.0483 & 0.1191 & \textcolor{best}{0.2018} & 0.1516 & \textcolor{second}{0.1635} \\
 & NMI & 0.3270 & 0.2084 & 0.3409 & \textcolor{best}{0.6287} & 0.3413 & \textcolor{second}{0.6227} \\
 & ARI & 0.0787 & 0.0449 & 0.0966 & \textcolor{second}{0.7466} & 0.0854 & \textcolor{best}{0.7641} \\
 & RI & 0.6166 & 0.4713 & 0.8420 & \textcolor{second}{0.8909} & 0.7598 & \textcolor{best}{0.8980} \\
\midrule
\multirow{4}{*}{Adiac} & F1 & \textcolor{second}{0.0353} & 0.0000 & 0.0315 & 0.0067 & 0.0226 & \textcolor{best}{0.0689} \\
 & NMI & 0.6162 & 0.3043 & 0.3845 & \textcolor{second}{0.6859} & 0.5294 & \textcolor{best}{0.7134} \\
 & ARI & 0.2292 & 0.0195 & 0.0294 & \textcolor{second}{0.9176} & 0.1317 & \textcolor{best}{0.9264} \\
 & RI & 0.9452 & 0.3472 & 0.9359 & \textcolor{second}{0.9677} & 0.9497 & \textcolor{best}{0.9713} \\
\midrule
\multirow{4}{*}{AllGestureWiimoteX} & F1 & 0.0685 & 0.0510 & 0.0074 & 0.0546 & \textcolor{second}{0.0888} & \textcolor{best}{0.1643} \\
 & NMI & 0.3577 & 0.0519 & 0.2404 & \textcolor{best}{0.4319} & 0.1468 & \textcolor{second}{0.4016} \\
 & ARI & 0.2466 & 0.0041 & 0.0457 & \textcolor{best}{0.5557} & 0.0485 & \textcolor{second}{0.4501} \\
 & RI & 0.8624 & 0.2975 & \textcolor{best}{0.8827} & \textcolor{second}{0.8696} & 0.7828 & 0.8250 \\
\midrule
\multirow{4}{*}{AllGestureWiimoteZ} & F1 & 0.0496 & 0.0610 & 0.0185 & \textcolor{second}{0.1395} & 0.0942 & \textcolor{best}{0.1654} \\
 & NMI & 0.2892 & 0.1058 & 0.1181 & \textcolor{best}{0.3696} & 0.1028 & \textcolor{second}{0.2901} \\
 & ARI & 0.1363 & 0.0419 & 0.0111 & \textcolor{best}{0.4720} & 0.0321 & \textcolor{second}{0.4417} \\
 & RI & 0.8246 & 0.6657 & \textcolor{best}{0.8751} & \textcolor{second}{0.8361} & 0.8060 & 0.8224 \\
\midrule
\multirow{4}{*}{BirdChicken} & F1 & \textcolor{second}{0.5000} & 0.4486 & 0.1067 & 0.3484 & 0.4949 & \textcolor{best}{0.8880} \\
 & NMI & 0.0000 & 0.0073 & \textcolor{second}{0.2982} & 0.0217 & 0.0000 & \textcolor{best}{0.6062} \\
 & ARI & 0.0556 & 0.0444 & \textcolor{second}{0.0870} & 0.0605 & 0.0532 & \textcolor{best}{0.6858} \\
 & RI & 0.4737 & 0.4789 & \textcolor{second}{0.5579} & 0.5500 & 0.4737 & \textcolor{best}{0.8577} \\
\midrule
\multirow{4}{*}{Car} & F1 & 0.3330 & 0.1343 & 0.1090 & \textcolor{best}{0.6081} & 0.1684 & \textcolor{second}{0.4166} \\
 & NMI & \textcolor{second}{0.2542} & 0.1787 & 0.0535 & \textcolor{best}{0.5935} & 0.0910 & 0.2219 \\
 & ARI & 0.1361 & 0.0877 & 0.0282 & \textcolor{best}{0.6727} & 0.0175 & \textcolor{second}{0.2845} \\
 & RI & 0.6271 & \textcolor{second}{0.6644} & 0.6322 & \textcolor{best}{0.8408} & 0.5181 & 0.6417 \\
\midrule
\multirow{4}{*}{CricketX} & F1 & \textcolor{best}{0.2002} & 0.0403 & 0.0128 & 0.0324 & 0.1212 & \textcolor{second}{0.1526} \\
 & NMI & \textcolor{best}{0.4642} & 0.0812 & 0.2403 & \textcolor{second}{0.4437} & 0.1212 & 0.3695 \\
 & ARI & 0.2472 & 0.0207 & 0.0294 & \textcolor{best}{0.7701} & 0.0204 & \textcolor{second}{0.7460} \\
 & RI & 0.8749 & 0.5700 & \textcolor{second}{0.8924} & \textcolor{best}{0.9026} & 0.8257 & 0.8913 \\
\midrule
\multirow{4}{*}{CricketZ} & F1 & 0.0709 & 0.0460 & 0.0236 & \textcolor{second}{0.0746} & 0.0668 & \textcolor{best}{0.1223} \\
 & NMI & 0.4557 & 0.0997 & 0.2450 & \textcolor{best}{0.4903} & 0.1568 & \textcolor{second}{0.4632} \\
 & ARI & 0.2339 & 0.0320 & 0.0424 & \textcolor{second}{0.7778} & 0.0402 & \textcolor{best}{0.7821} \\
 & RI & 0.8778 & 0.6671 & 0.8874 & \textcolor{second}{0.9060} & 0.8366 & \textcolor{best}{0.9082} \\
\midrule
\multirow{4}{*}{Crop} & F1 & \textcolor{second}{0.0388} & 0.0044 & 0.0003 & 0.0228 & 0.0387 & \textcolor{best}{0.0474} \\
 & NMI & 0.4101 & 0.3851 & 0.2841 & \textcolor{second}{0.5367} & 0.4110 & \textcolor{best}{0.5374} \\
 & ARI & 0.2295 & 0.1547 & 0.0172 & \textcolor{second}{0.6781} & 0.1816 & \textcolor{best}{0.7096} \\
 & RI & 0.9271 & 0.8087 & \textcolor{best}{0.9568} & 0.9252 & 0.9308 & \textcolor{second}{0.9335} \\
\midrule
\multirow{4}{*}{DistalPhalanx} & F1 & 0.3016 & 0.0906 & 0.0090 & \textcolor{second}{0.3824} & 0.1637 & \textcolor{best}{0.4335} \\
 & NMI & \textcolor{second}{0.3647} & 0.2418 & 0.1551 & 0.3313 & 0.1277 & \textcolor{best}{0.4556} \\
 & ARI & 0.2625 & 0.1983 & 0.0842 & \textcolor{second}{0.4485} & 0.1188 & \textcolor{best}{0.6231} \\
 & RI & 0.6537 & 0.5878 & 0.6102 & \textcolor{second}{0.7586} & 0.5812 & \textcolor{best}{0.8286} \\
\midrule
\multirow{4}{*}{ECG200} & F1 & 0.5062 & \textcolor{second}{0.7455} & 0.0043 & 0.3850 & 0.3200 & \textcolor{best}{0.8354} \\
 & NMI & 0.0044 & \textcolor{second}{0.1885} & 0.1304 & 0.0498 & 0.0707 & \textcolor{best}{0.3635} \\
 & ARI & 0.0170 & \textcolor{second}{0.2791} & 0.0397 & 0.1147 & 0.1166 & \textcolor{best}{0.4950} \\
 & RI & 0.5240 & \textcolor{second}{0.6422} & 0.4933 & 0.5739 & 0.5604 & \textcolor{best}{0.7508} \\
\midrule
\multirow{4}{*}{ElectricDevices} & F1 & 0.1145 & 0.1229 & 0.0227 & \textcolor{second}{0.1543} & 0.1026 & \textcolor{best}{0.2070} \\
 & NMI & \textcolor{second}{0.3952} & 0.1421 & 0.0000 & \textcolor{best}{0.5630} & 0.1579 & 0.3887 \\
 & ARI & 0.3169 & 0.0456 & 0.0000 & \textcolor{best}{0.7791} & 0.0770 & \textcolor{second}{0.5721} \\
 & RI & \textcolor{second}{0.7979} & 0.4529 & 0.1774 & \textcolor{best}{0.8987} & 0.7306 & 0.7860 \\
\midrule
\multirow{4}{*}{FaceAll} & F1 & 0.0820 & 0.0413 & 0.0043 & 0.0407 & \textcolor{second}{0.0892} & \textcolor{best}{0.1076} \\
 & NMI & \textcolor{best}{0.6771} & 0.3615 & 0.1204 & \textcolor{second}{0.5181} & 0.1705 & 0.3063 \\
 & ARI & \textcolor{best}{0.5839} & 0.1676 & 0.0090 & \textcolor{second}{0.4570} & 0.0683 & 0.4295 \\
 & RI & \textcolor{best}{0.9281} & 0.7767 & \textcolor{second}{0.8925} & 0.8426 & 0.8372 & 0.8542 \\
\midrule
\multirow{4}{*}{FreezerRegularTrain} & F1 & \textcolor{best}{0.7592} & 0.0837 & 0.0009 & 0.2735 & 0.5714 & \textcolor{second}{0.7209} \\
 & NMI & \textcolor{best}{0.2070} & 0.1430 & 0.1199 & 0.0887 & 0.0902 & \textcolor{second}{0.1726} \\
 & ARI & \textcolor{best}{0.2694} & 0.1361 & 0.0160 & 0.1175 & 0.0544 & \textcolor{second}{0.1986} \\
 & RI & \textcolor{best}{0.6347} & 0.5681 & 0.5082 & 0.5588 & 0.5271 & \textcolor{second}{0.5993} \\
\midrule
\multirow{4}{*}{SyntheticControl} & F1 & \textcolor{second}{0.1603} & 0.0303 & 0.0286 & 0.0738 & 0.0650 & \textcolor{best}{0.3153} \\
 & NMI & \textcolor{best}{0.9475} & 0.6563 & 0.1059 & \textcolor{second}{0.7297} & 0.4849 & 0.4635 \\
 & ARI & \textcolor{best}{0.9376} & 0.4811 & 0.0144 & \textcolor{second}{0.6795} & 0.3551 & 0.5933 \\
 & RI & \textcolor{best}{0.9829} & 0.7971 & 0.7998 & \textcolor{second}{0.8530} & 0.8142 & 0.8117 \\
\midrule
\textbf{Global} & \#Best & 12 & 0 & 3 & 16 & 0 & 29 \\
 & \#Second & 8 & 5 & 5 & 23 & 2 & 17 \\
 & Avg Rank & 2.98 & 4.68 & 4.67 & \textcolor{second}{2.47} & 4.35 & \textcolor{best}{1.85} \\
\end{tabular}%
}
\label{tab:benchmark}
\end{table}

\subsection{Comparison Methods}
The proposed approach has been compared against a set of representative baselines for time series clustering.
\begin{itemize}
    \item \textbf{TimeSeriesKMeans-DTW (TSK-DTW)} \cite{dtw} is a classical clustering method that extends K-means using Dynamic Time Warping as similarity measure, providing a strong non-deep baseline.
    \item \textbf{HDBSCAN} \cite{hdbscan} is a hierarchical density-based clustering algorithm that extends DBSCAN by constructing a hierarchy of clusters, enabling robust clustering with varying densities and improved handling of noise.
    \item \textbf{FCACC} \cite{FCACC} is a deep clustering framework that integrates contrastive learning with fuzzy clustering, enabling soft cluster assignments in the learned latent space.
    \item \textbf{PG-Mamba} \cite{pgmamba} is a recent model that combines Mamba-based sequence modeling with a graph-based clustering framework, leveraging relational structures between time series.
    \item \textbf{EMTC} \cite{emtc} is a multi-view self-supervised approach that employs masking strategies and contrastive learning to learn robust representations for time series clustering.
\end{itemize}

\subsection{Evaluation Metrics}

The clustering performance is evaluated using the following standard metrics:

\begin{itemize}
    \item \textbf{NMI (Normalized Mutual Information):} measures the shared information between predicted clusters and ground-truth labels.
    
    \item \textbf{RI (Rand Index):} evaluates the pairwise agreement between predicted and true cluster assignments.
    
    \item \textbf{ARI (Adjusted Rand Index):} measures the similarity between clusterings while correcting for chance.
    
    \item \textbf{F1-score:} represents the harmonic mean of precision and recall.
\end{itemize}

All metrics range between 0 and 1, where higher values indicate better clustering performance.

\subsection{Experimental Details}
For the baseline methods, we implement them using open-source code, following the parameter configurations and experimental setups specified in their respective papers. \\
For the proposed FMMVCC model, the specific parameter settings are $\lambda_b = 0.2$ and $\lambda_s = 0.5$. These values were selected based on a preliminary sensitivity analysis to ensure an optimal balance between cluster separation and assignment stability, as further described in Section \ref{sec:parameter}. The batch size is set to 32, the latent representation dimension is fixed to 64, the drop rate $\rho$ is set to 0.3, the maximal length $L_{max}$ is assigned to 5, and the number of views $N$ is equal to 4, as described in Section \ref{sec:views}. The model is trained using the AdamW optimizer with an initial learning rate of 1e-3 during the pretraining phase and 1e-4 during the finetuning phase. The number of pretraining epochs and finetuning stage is set to 100. \\
All training and evaluation procedures were performed using a NVIDIA Quadro RTX 8000.

\begin{figure}[!t]
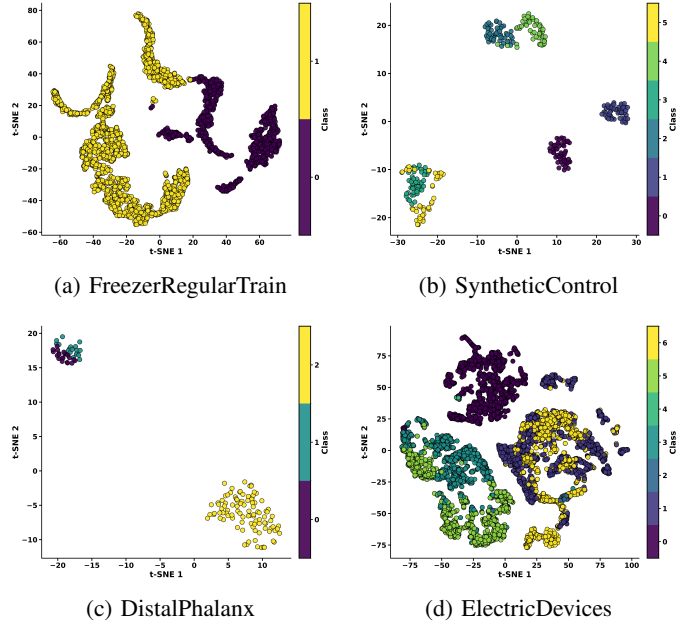

    \centering
    \begin{subfigure}[b]{0.48\columnwidth}
        \centering
        \includegraphics[width=\linewidth]{images/Train.png}
        \caption{FreezerRegularTrain}
    \end{subfigure}
    \hfill
    \begin{subfigure}[b]{0.48\columnwidth}
        \centering
        \includegraphics[width=\linewidth]{images/Synthetic.png}
        \caption{SyntheticControl}
    \end{subfigure}

    \vspace{0.3cm}

    \begin{subfigure}[b]{0.48\columnwidth}
        \centering
        \includegraphics[width=\linewidth]{images/Phalanx.png}
        \caption{DistalPhalanx}
    \end{subfigure}
    \hfill
    \begin{subfigure}[b]{0.48\columnwidth}
        \centering
        \includegraphics[width=\linewidth]{images/ElectricDevices.png}
        \caption{ElectricDevices}
    \end{subfigure}

    \caption{Latent space visualizations of different datasets.}
    \label{fig:latent_spaces}
\end{figure}

\subsection{Performance Comparing}
Table \ref{tab:benchmark} reports the clustering results across all datasets, where the best performances are highlighted in red and the second-best in blue. In general, FMMVCC achieves the highest number of best scores (29) and the lowest average per-metric rankings across all datasets (1.85), demonstrating a clear and consistent advantage over all baselines. \\
This advantage is particularly evident in structure-aware metrics such as ARI and NMI, where FMMVCC often ranks first or second. This suggests that the learned latent space better captures the underlying cluster structure, rather than simply optimizing pairwise agreements (as reflected by RI). In contrast, some baselines achieve competitive RI values but significantly lower ARI, indicating less meaningful cluster assignments. \\
Compared to other deep learning approaches, FMMVCC shows a clear improvement in robustness and stability. Although methods such as FCACC achieve top performance in some cases, their results are more variable, as reflected by a higher average rank (2.47). Similarly, EMTC and PG-Mamba exhibit less consistent behavior across metrics, suggesting that masking or Mamba-based modeling alone is not sufficient without a tightly integrated multi-view contrastive framework. \\
Traditional methods such as TSK-DTW and HDBSCAN generally obtain lower average rankings; however, they remain competitive in specific scenarios, occasionally achieving strong results, particularly in pairwise-based metrics or when the cluster structure is well aligned with the underlying similarity measure. This highlights that classical approaches can still be effective in simpler settings, but lack the representational flexibility required to consistently capture complex temporal dependencies across diverse datasets. \\
Overall, the results indicate that combining contrastive multi-view learning, temporal masking, and the causal selective dynamics of Mamba leads to a more stable and structurally consistent latent space, resulting in improved clustering performance across a wide range of scenarios. Some examples of the learned latent spaces across different datasets are illustrated in Figure \ref{fig:latent_spaces}.

\begin{figure}[!t]
    \centering
    \includegraphics[width=\columnwidth]{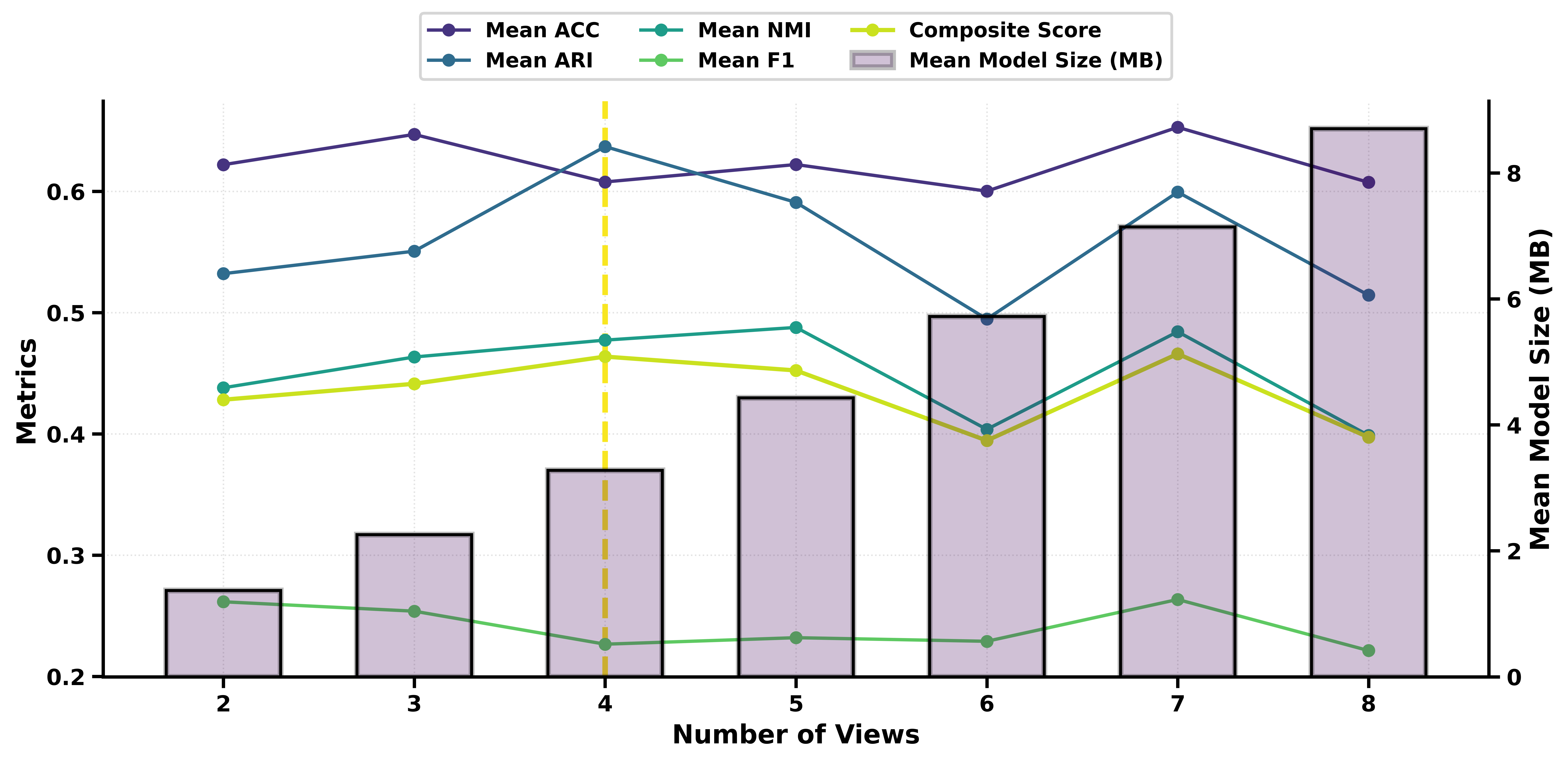}
    \caption{Clustering performance and model size as a function of the number of views $N$. The dashed yellow line indicates $N=4$, representing the selected parameter.}
    \label{fig:trade-off}
\end{figure}

\subsection{Sensitivity to the Number of Views}\label{sec:views}
To assess the impact of the multi-view strategy, we vary $N$ from 2 to 8, across the whole time-series datasets. Performance is monitored via standard metrics and a Composite Score, defined as the arithmetic mean of the other scores, providing a holistic evaluation that balances structural and pairwise consistency. \\
As shown in Fig. \ref{fig:trade-off}, the Composite Score and ARI exhibit a non-monotonic trend with an optimal peak at $N=4$. While $N=7$ slightly improves ACC, it increases the memory footprint from 3.5 MB to over 6 MB, whereas $N=8$ leads to performance degradation due to redundant noise. The stability of F1 and NMI between $N=3$ and $N=5$ confirms the robustness of FMMVCC, validating $N=4$ (dashed yellow line) as the ideal trade-off between representation richness and computational efficiency.

\subsection{Clustering Loss} 
\begin{figure}[!b]
    \centering
    \includegraphics[width=\columnwidth]{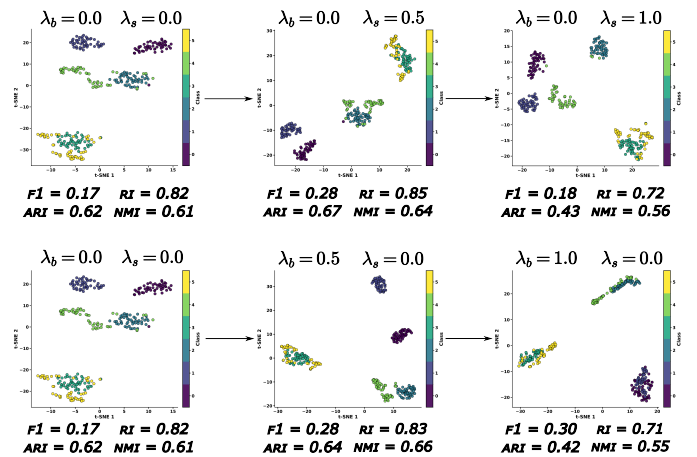}
    \caption{Impact of loss weights on cluster formation. The top row depicts how the separation term $\lambda_s$ enhances cluster isolation by maximizing inter-centroid distances. The bottom row demonstrates the influence of the balance term $\lambda_b$ on cluster distribution uniformity.}
    \label{fig:parameters}
\end{figure}

Analysis\label{sec:parameter}
To evaluate the effect of fuzzy clustering loss components, a sensitivity analysis is conducted on the \textit{SyntheticControl} dataset by varying the separation weight $\lambda_s$ and the balance weight $\lambda_b$. Fig. \ref{fig:parameters} shows the corresponding t-SNE projections \cite{tsne}. \\
The first row illustrates the impact of $\lambda_s \in \{0, 0.5, 1.0\}$ with $\lambda_b = 0$. When $\lambda_s = 0$, latent representations exhibit partial overlap, making the boundaries of the cluster less distinguishable. As $\lambda_s$ increases, clusters become progressively more separated, with clearer margins between groups. This indicates that the separation term effectively enhances the inter-cluster distance and improves the discriminability of the learned representations. \\
The second row shows the effect of $\lambda_b \in \{0, 0.5, 1.0\}$ with $\lambda_s = 0$. In this case, increasing $\lambda_b$ leads to more compact and structured clusters, with samples concentrating more tightly around their respective centers. This suggests that the balance term encourages a more uniform utilization of clusters and reduces ambiguity in the assignments, resulting in lower intra-cluster variance. \\
Overall, the results highlight the complementary roles of the two components: $\lambda_s$ primarily controls inter-cluster separation, while $\lambda_b$ improves intra-cluster compactness.

\subsection{Scalability Analysis}\label{sec:scalability}
\begin{figure}[!t]
    \centering
    \includegraphics[width=\columnwidth]{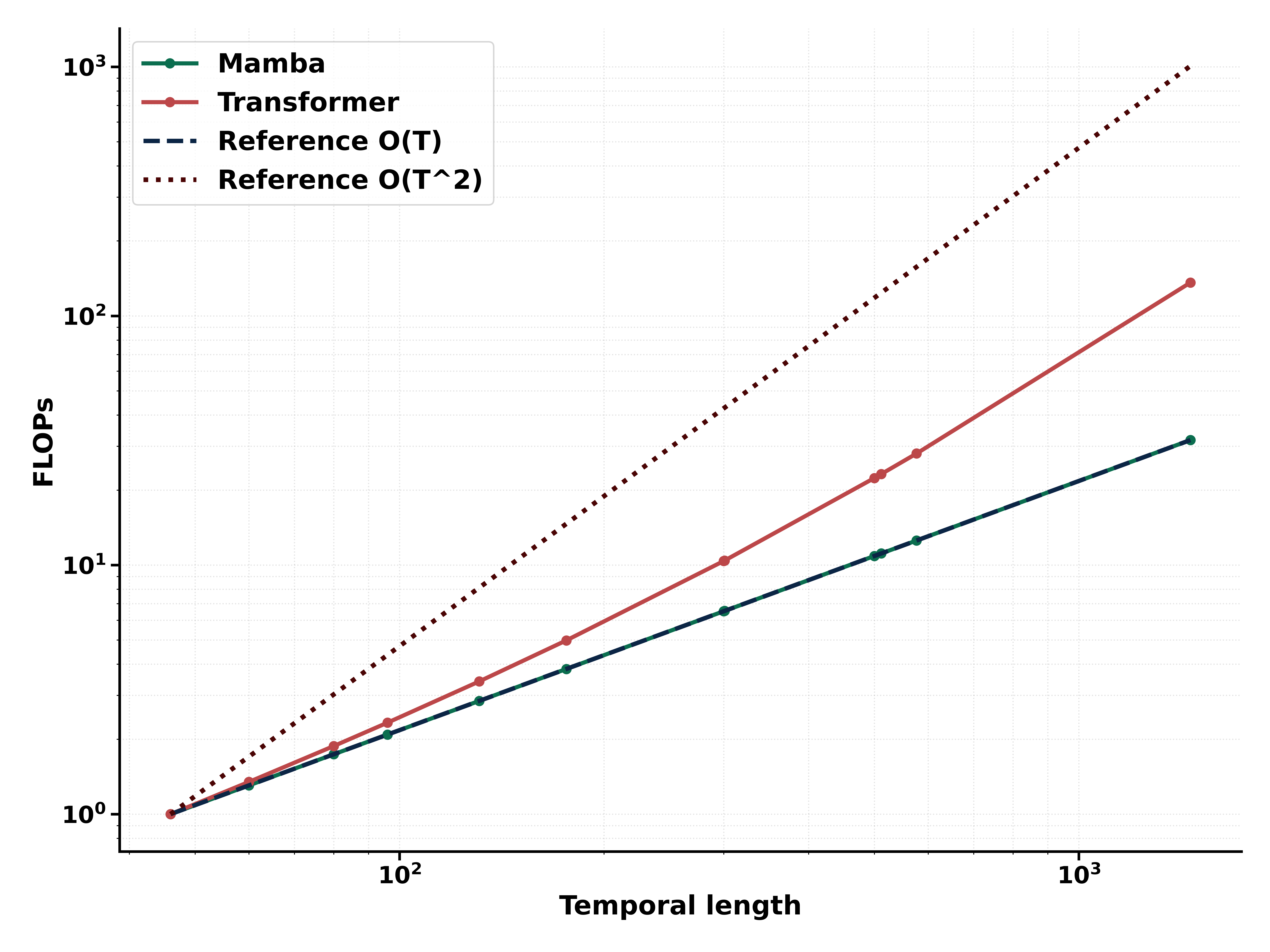}
    \caption{Scalability plot of FLOPs with respect to sequence length $T$ for Mamba and Transformer encoders in a log-log scale, with $O(T)$ and $O(T^2)$ reference trends.}
    \label{fig:scalability}
\end{figure}

To validate the computational efficiency of the proposed Mamba-based encoder, scalability is analyzed in terms of floating point operations (FLOPs) as a function of the sequence length $T$. For a fair comparison, the Mamba encoder is replaced with a Transformer-based encoder while keeping the rest of the framework unchanged, including architecture, training setup, and hyperparameters. The Transformer is configured with a single attention head to avoid introducing additional capacity bias. \\
Figure \ref{fig:scalability} shows the FLOPs growth together with reference $O(T)$ and $O(T^2)$ trends. The Mamba encoder follows a linear scaling behaviour, while the Transformer exhibits a significantly steeper increase consistent closer to the quadratic complexity. The gap becomes more pronounces as $T$ increases, highlighting the superior scalability of the proposed approach. \\
These results confirm that the proposed framework preserves the linear-time complexity of Mamba, making it more suitable for long time series and large-scale scenarios.

\subsection{Ablation Study}
To assess the impact of the sequence modeling strategy, we compare the unidirectional Mamba encoder with a bidirectional variant, where forward and backward representations are combined through a learnable gate \cite{bimamba}. \\
As shown in Table \ref{tab:metric_degradation}, the bidirectional configuration consistently degrades performance across all metrics. The largest drops are observed in ARI ($-12.04\%$) and NMI ($-11.57\%$), indicating a weaker alignment with the structure of the ground-truth, followed by F1 ($-9.41\%$). RI shows a smaller decrease ($-3.64\%$), but still confirms the overall trend. \\
This behavior suggests that enforcing a causal, unidirectional representation is more suitable for the proposed framework. In particular, the bidirectional setting may introduce temporal inconsistencies across masked views, reducing the effectiveness of the contrastive objective and leading to less structured latent representations.

\begin{table}[!t]
\centering
\small
\setlength{\tabcolsep}{4pt}
\caption{Per-metric analysis of Bidirectional vs.\ Unidirectional Mamba Encoders.}
\label{tab:metric_degradation}
\begin{tabular}{c|c|c|c}
\toprule
Avg Metric & Unidirectional & Bidirectional & $\Delta$\,\% \\
\midrule
F1  & 0.3206 & 0.2873 & $\downarrow$ 9.41\% \\
NMI & 0.4251 & 0.3814 & $\downarrow$ 11.57\% \\
ARI & 0.5801 & 0.5245 & $\downarrow$ 12.04\% \\
RI  & 0.8253 & 0.7954 & $\downarrow$ 3.64\% \\
\bottomrule
\end{tabular}
\end{table}

\section{Conclusions}\label{sec:conclusions}
This paper presents an innovative deep clustering framework for univariate time series, integrating Mamba with a multi-view contrastive learning strategy. Our model effectively captures long-range temporal dependencies by taking advantage of the Mamba architecture's linear complexity. The addition of a stochastic temporal masking and augmentation strategy ensures that the learned representations can handle noise and missing data, which are common problems in real-world IoT and sensor environments. The addition of a fuzzy clustering objective also makes it possible to assign soft clusters, which makes it easier to deal with unclear temporal patterns. \\
Experimental evaluations conducted on several benchmark datasets from the UCR Archive show that FMMVCC consistently outperforms the baseline methods. Our analysis confirms that a unidirectional encoding strategy, in conjunction with inter-view and intra-view reconstruction losses, is optimal for preserving temporal consistency in a self-supervised framework. The parameter sensitivity study underscores the significance of the separation and balance terms in the configuration of a discriminative latent space. \\
Despite these results, the current framework is primarily designed for univariate sequences, and its latent representations lack direct interpretability. Therefore, future research will focus on extending the framework to multivariate time series to capture inter-variable correlations, alongside conducting a deeper latent space analysis to enhance the explainability of the learned temporal features and their influence on cluster formation.

\section*{Data and Resources}
The data used in this study are in \href {https://www.cs.ucr.edu/%7Eeamonn/time_series_data_2018/} {UCR Time Series Classification Archive}. The implementation of FMMVCC is available on \href{https://github.com/DonatoCerciello/FMMVCC}{https://github.com/DonatoCerciello/FMMVCC}.
\section*{Acknowledgment}

This work was supported by the MSCA Doctoral Networks project T.U.A.I. -  Towards an Understanding of Artificial Intelligence via a transparent, open and explainable perspective (TUAI) project, N°101168344.

\bibliographystyle{IEEEtran}
\bibliography{bibliography}
\end{document}